%% file: main.tex
\PassOptionsToPackage{table,xcdraw,dvipsnames}{xcolor}
\documentclass[10pt,twocolumn,letterpaper]{article}

\usepackage{cvpr}
\input{preamble}

\input{command}

\title{Embracing Collaboration Over Competition:\\ Condensing Multiple Prompts for Visual In-Context Learning}

\author{
Jinpeng Wang$^{1}\thanks{These authors contributed equally to this work.}$ , \ 
Tianci Luo$^{2}{}^{*}$, \ 
Yaohua Zha$^{1}$, \ 
Yan Feng$^{6}$, \ 
Ruisheng Luo$^{1}$, \\
Bin Chen$^{2}\thanks{Corresponding author.}$ , \ 
Tao Dai$^{3}$, \ 
Long Chen$^{4}$, \ 
Yaowei Wang$^{5}$, \ 
Shu-Tao Xia$^{1,5}$ \\
{\small $^1$Tsinghua Shenzhen International Graduate School, Tsinghua University} \\
{\small $^2$Harbin Institute of Technology, Shenzhen} \\
{\small $^3$Shenzhen University \quad 
$^4$The Hong Kong University of Science and Technology} \\
{\small $^5$Research Center of Artificial Intelligence, Peng Cheng Laboratory \quad
$^6$Meituan, Beijing} \\
{\small \tt wjp20@mails.tsinghua.edu.cn \qquad
 \Letter\ \tt chenbin2021@hit.edu.cn} \\
}

\begin{document}
\maketitle
\input{sections/Abstract}    
\input{sections/Introduction}
\input{sections/RelatedWorks}
\input{sections/Method}
\input{sections/Experiments}
\input{sections/Conclusions}
{
    \small
    \bibliographystyle{ieeenat_fullname}
    \bibliography{main}
}

\end{document}

%% file: preamble.tex
\usepackage{graphicx}
\usepackage{amsmath}
\usepackage{amssymb}
\usepackage{booktabs}
\usepackage[misc]{ifsym}

\definecolor{cvprblue}{rgb}{0.21,0.49,0.74}
\usepackage[pagebackref,breaklinks,colorlinks,allcolors=cvprblue]{hyperref}

\usepackage{url}
\usepackage{amsfonts}
\usepackage{nicefrac}
\usepackage{microtype}

\usepackage{makecell}
\usepackage{amsthm}
\usepackage{bm}
\usepackage{multicol}
\usepackage{colortbl}
\usepackage{multirow}
\usepackage{enumitem}
\usepackage{bbding}
\usepackage{color}

\usepackage{algorithm}
\usepackage{listings}
\usepackage{natbib}

\usepackage{etoolbox}
\makeatletter
\AfterEndEnvironment{algorithm}{\let\@algcomment\relax}
\AtEndEnvironment{algorithm}{\kern2pt\hrule\relax\vskip3pt\@algcomment}
\let\@algcomment\relax
\newcommand\algcomment[1]{\def\@algcomment{\footnotesize#1}}
\renewcommand\fs@ruled{\def\@fs@cfont{\bfseries}\let\@fs@capt\floatc@ruled
  \def\@fs@pre{\hrule height.8pt depth0pt \kern2pt}%
  \def\@fs@post{}%
  \def\@fs@mid{\kern2pt\hrule\kern2pt}%
  \let\@fs@iftopcapt\iftrue}
\makeatother

%% file: command.tex
\renewcommand{\paragraph}[1]{\medskip\noindent\textbf{#1.~}}

\newcommand{\xiaok}[1]{\left(#1\right)}
\newcommand{\zhongk}[1]{\left[#1\right]}
\newcommand{\dak}[1]{\left\{#1\right\}}

\newcommand{\bmA}{\bm{A}}

\newcommand{\bmF}{\bm{F}}

\newcommand{\bmT}{\bm{T}}

\newcommand{\bmW}{\bm{W}}

\newcommand{\calL}{\mathcal{L}}

\newcommand{\calP}{\mathcal{P}}
\newcommand{\calQ}{\mathcal{Q}}

\newcommand{\calS}{\mathcal{S}}

\newcommand{\bbE}{\mathbb{E}}

\newcommand{\bbR}{\mathbb{R}}

\newcommand{\modelname}{\textsc{Condenser}}

\definecolor{myblue}{HTML}{053CF5}
\definecolor{myorange}{HTML}{ED6E57}
\definecolor{mygreen}{HTML}{81D453}

%% file: sections/Abstract.tex
\begin{abstract}
Visual In-Context Learning (VICL) enables adaptively solving vision tasks by leveraging pixel demonstrations, mimicking human-like task completion through analogy. 
Prompt selection is critical in VICL, but current methods assume the existence of a single ``ideal" prompt in a pool of candidates, which in practice may not hold true. 
Multiple suitable prompts may exist, but individually they often fall short, leading to difficulties in selection and the exclusion of useful context. 
To address this, we propose a new perspective: \textbf{prompt condensation}. Rather than relying on a single prompt, candidate prompts collaborate to efficiently integrate informative contexts without sacrificing resolution. 
We devise \modelname{}, a lightweight external plugin that compresses relevant fine-grained context across multiple prompts. 
Optimized end-to-end with the backbone, \modelname{} ensures accurate integration of contextual cues. 
Experiments demonstrate \modelname{} outperforms state-of-the-arts across benchmark tasks, showing superior context compression, scalability with more prompts, and enhanced computational efficiency compared to ensemble methods, positioning it as a highly competitive solution for VICL.  
Code is open-sourced at \url{https://github.com/gimpong/CVPR25-Condenser}.
\end{abstract}

%% file: sections/Introduction.tex
\section{Introduction}
\label{sec:intro}

\begin{figure}[t]
    \centering
    \includegraphics[width=\columnwidth]{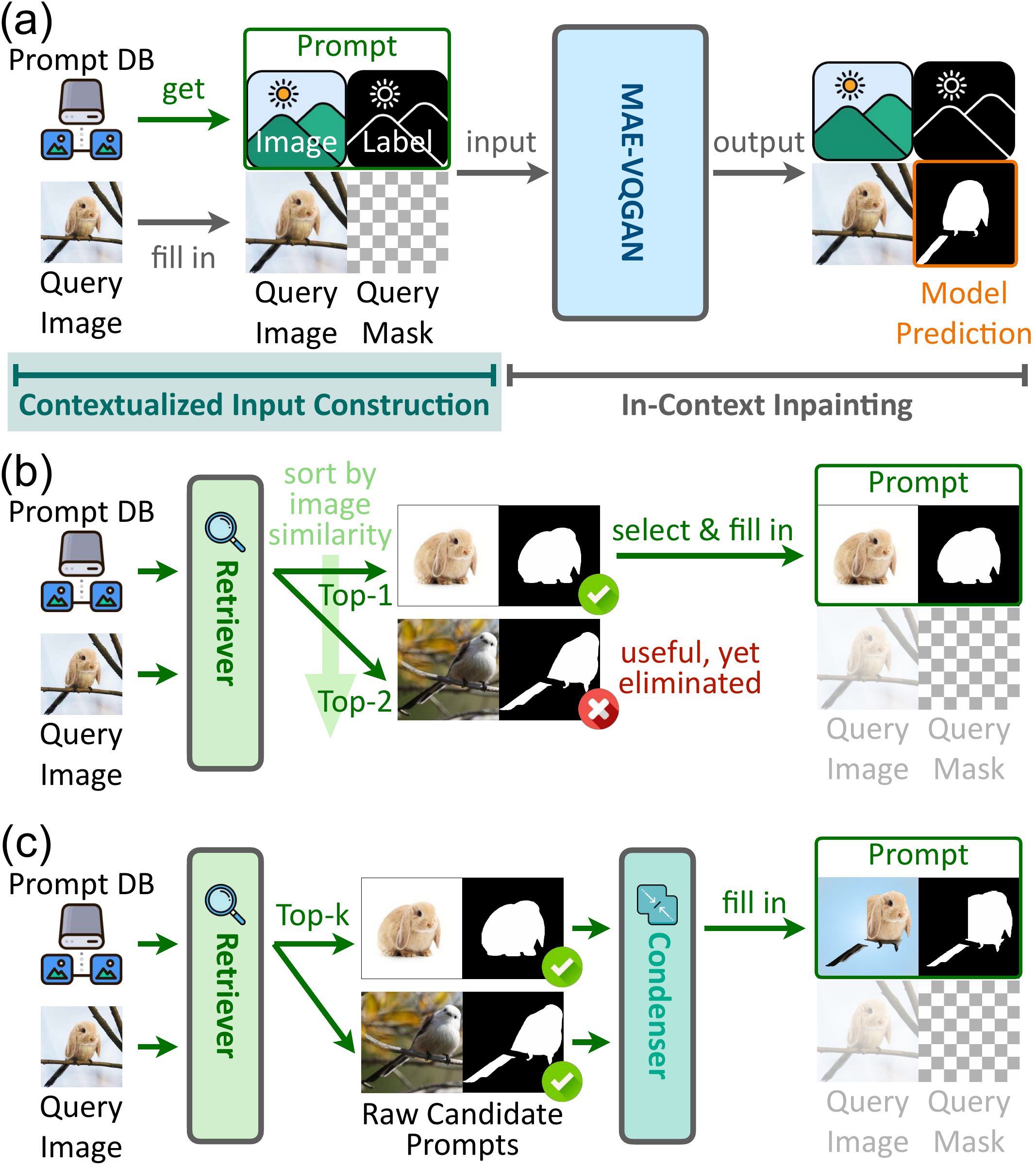}
    \caption{(\textbf{a}) The impainting framework \cite{bar2022visual} is a typical prototype for Visual In-Context Learning (VICL), where a fixed-size input canvas is divided into regions. 
    The prompt image and its label occupy the top half, while the query image is placed in the bottom left. 
    The vision backbone is tasked with recovering the bottom right region, which represents the query label, by reasoning within the pixel context. 
    Prompt selection plays a crucial role in constructing the input and significantly impacts model predictions.
    (\textbf{b}) Most existing VICL methods view prompt selection as a competitive process, where a single prompt is chosen, often discarding useful alternatives. Meanwhile, the top-1 prompt may not be ideal for the query image.
    (\textbf{c}) We introduce the concept of \emph{collaborative condensation}, where multiple raw prompts are combined to leverage their individual strengths into a single condensed prompt. The proposed \modelname{} is a lightweight, flexible, and efficient plugin to the vision backbone, effective across a range of VICL tasks.}
    \label{fig:intro}
\end{figure}

Humans can complete tasks through imitation and analogy, even when these tasks are difficult or unfamiliar. 
\emph{Could a similar paradigm be extended to machines?}
This fundamental question has evoked research on In-Context Learning (ICL) \cite{ICL_Survey_2024}, especially with the advent of foundational models \cite{brown2020language,team2023gemini,touvron2023llama}. 
Recent work with Large Language Models (LLMs) \cite{hendel2023context,wei2023larger} and Multimodal LLMs (MLLMs) \cite{Mimic-it_2023,MMICL_2024,zhou2024visual} has demonstrated that prompting models using task-specific examples can efficiently adapt downstream tasks, achieving comparable efficacy to in-weight learning \cite{dai2023can}. 
More inspiringly, the vision community has revealed that vision backbones \cite{bar2022visual,wang2023seggpt} can also exhibit this property, unlocking a novel paradigm as Visual ICL (VICL) \cite{wang2023seggpt,fang2024explore,li2024visual}.
As shown in \Cref{fig:intro}(a), VICL enables leveraging contextual pixel cues from prompts (\ie, demonstrations), which helps vision backbones solve tasks adaptively and effectively. 

Prompt selection plays a crucial role in VICL. 
While the early work \cite{bar2022visual} selected prompts randomly, \citet{zhang2024makes} suggested choosing prompts with higher relevance can notably enhance effectiveness. 
Based on this insight, a range of efforts \cite{zhang2024makes,sun2023exploring,xu2024towards} has been devoted to selecting the \emph{best} prompt for the query image, typically using backbone's task performance conditioned by single prompts as signals to learn prompt retrieval or ranking. 
Albeit remarkable progress, these methods are soon reaching a plateau. 
Essentially, they operate on a competitive assumption—that there exists a \emph{perfect} prompt within the database to be retrieved. 
However, as illustrated in \Cref{fig:intro}(b), this assumption may be challenged in practice. 
There are often multiple suitable prompts, but each one alone may fall short of being \emph{ideal}. 
In such cases, prompt selection may struggle to make a single choice, inevitably dropping helpful prompts as context. 
This limitation could restrict the generalizability of VICL.

We believe prompt candidates necessity \emph{collaboration} rather than \emph{exclusive competition} in VICL. 
Combining multiple prompts is expected to enhance the compositional generalization ability, yet preliminary solutions, including downsampling and ensemble, still have clear limitations.
The downsampling methods \cite{bar2022visual,wang2023images,zhang2024makes} scale down sub-images proportionally to arrange more prompts on the fixed-size input canvas, while they may lose visual details in prompts. 
Ensemble methods \cite{sun2023exploring,xu2024towards}, on the other hand, separately combine each candidate prompt with the query image to create multiple inputs and finally aggregate their contextual outputs. 
Output aggregation is quite tricky for specific tasks and often requires careful adjustments. What's worse, it raises efficiency concerns as the candidate prompts increase. 
Beyond these practices, we introduce a new research perspective of \textbf{condensation}, aiming to \emph{efficiently} integrate informative contexts in different raw prompts \emph{without sacrificing resolution} for better VICL.

We propose an end-to-end optimized approach to condense multiple prompts, as demonstrated in \Cref{fig:intro}(c). 
Given a set of candidate prompts retrieved along with the query image, we devise a lightweight \modelname{} to adaptively fuse informative cues. 
The \modelname{} identifies relevant image patterns and their associated annotations within the candidate prompts to form the reference for each query image patch, helping to extract the most relevant \emph{fine-grained} context in the final prompt.
We train \modelname{} jointly with the vision backbone, utilizing end-to-end feedback to optimize prompt condensation. 

Similar to retrievers, \modelname{} can act as a flexible external plugin to the vision backbone, making it highly adaptable and effective across VICL tasks. 
We conduct extensive experiments on several benchmark tasks, including foreground segmentation, single-object detection, and image coloring. 
Under standard protocols, even when using only a single candidate prompt, \modelname{} outperforms state-of-the-art methods remarkably, highlighting its strong information extraction ability. 
More encouragingly, \modelname{} performs better as the number of candidate prompts increases, further demonstrating its information integration capability. 
At the same time, it maintains a clear computational efficiency advantage over ensemble-based multi-prompt methods \cite{sun2023exploring}. 

To summarize, we make the following contributions.
\setlist{nolistsep}
\begin{itemize}[leftmargin=1.5em]
\item We successfully shift the paradigm of prompt selection in VICL from exclusive competition to collaborative condensation, which enables multiple prompts to work together effectively and efficiently. 
\item We design a lightweight \modelname{} to adaptively condense multiple prompts into a single, fine-grained prompt, enhancing VICL without sacrificing resolution.
\item Extensive experiments demonstrate that \modelname{} outperforms state-of-the-art methods on multiple tasks, showing effective context compression capacity, scalability with increasing candidate prompts, and favorable computational efficiency beyond output ensemble.
\end{itemize}

%% file: sections/RelatedWorks.tex
\begin{figure*}[t]
    \centering
    \includegraphics[width=\textwidth]{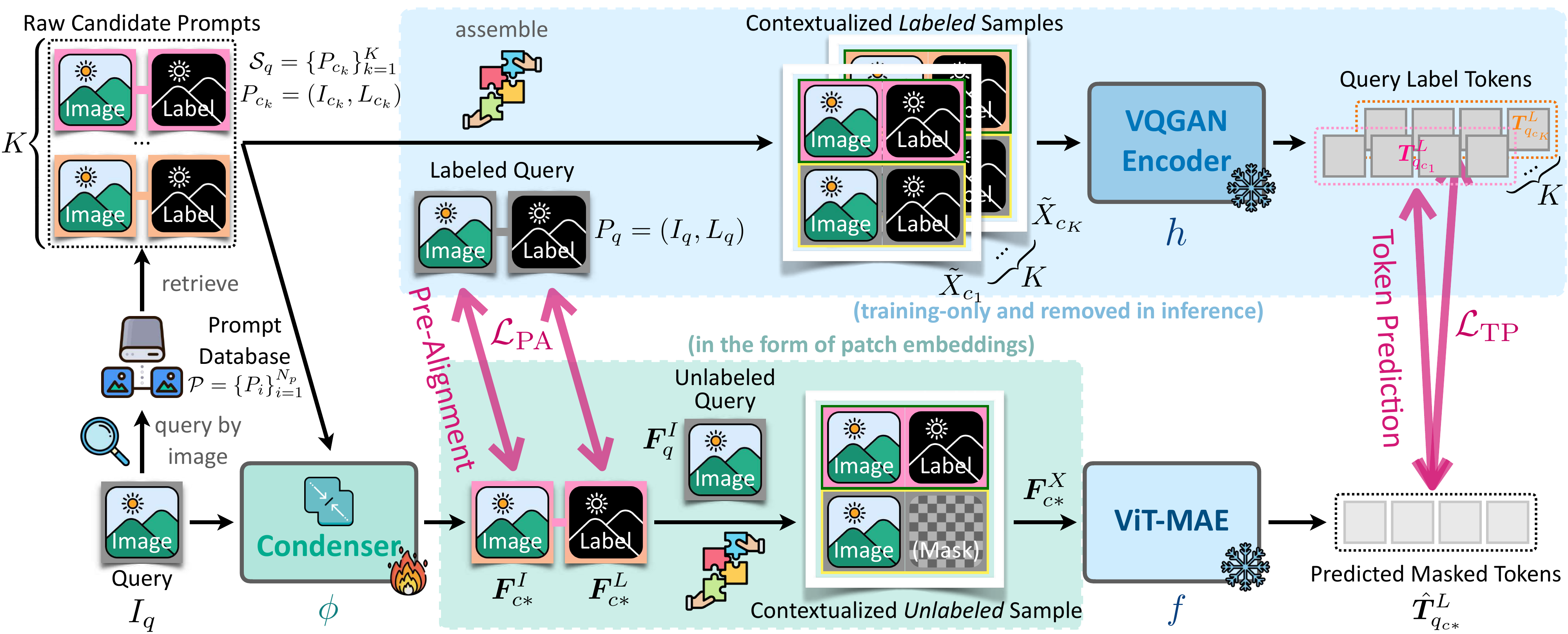}
    \caption{Prompt condensation with \modelname{} for VICL. 
    Given a query image $I_q$ from the training set, we first retrieve a set of candidate prompts $\calS_q$ from the prompt database $\calP$. 
    Then, we assemble each prompt from $\calS_q$ with the labeled query pair $P_q=(I_q,L_q)$ and construct $K$ contextualized labeled samples, namely $\tilde{X}_{c_1},\cdots,\tilde{X}_{c_K}$. 
    We encode each of them with a VQGAN encoder \cite{VQGAN_2021} $h(\cdot)$ and obtain a serial of query label tokens, $\bmT_{q_{c_1}}^L,\cdots,\bmT_{q_{c_K}}^L$ as signals. 
    Next, we transform and condense raw candidate prompts $\calS_q$ into an integrated prompt, $(\bmF_{c*}^I,\bmF_{c*}^L)$, using the proposed \modelname{}.
    The patches of condensed prompt and the query image form the contextualized unlabeled sample $\bmF_{c*}^X$, which is subsequently processed by the vision backbone $f(\cdot)$ to predict masked tokens $\hat{\bmT}_{q_{c_*}}^L$. 
    Finally, we incorporate token prediction and pre-alignment between the condensed prompt and the labeled query to guide optimization.}
    \label{fig:arc}
\end{figure*}

\section{Related Works}
\label{sec:related_works}
\subsection{In-Context Learning}
Large Language Models (LLMs) \cite{touvron2023llama,team2023gemini} have demonstrated outstanding ability and generalizability. 
The pioneering GPT-3 \cite{brown2020language} preliminary revealed LLMs' inherent capability to answer novel questions by mimicking responses from a set of input example pairs, which has triggered extensive explorations of In-Context Learning (ICL) in various applications \cite{press2022measuring,zheng2023can}. 
At the same time, many efforts \cite{why_work_1,why_work_2,why_work_3,why_work_4} have been devoted to the theoretical insights behind ICL, which supported its rationality and effectiveness.
The success of ICL has also paved the way into the multi-modal domain, including but not limited to multi-modal language models \cite{alayrac2022flamingo,peng2023kosmos,sun2024generative} and multimedia content generation \cite{diffusion,wang2023context,najdenkoska2024context}.

\subsection{Visual In-Context Learning} \label{subsec:icl}
Inspired by ICL's success in language processing and multimedia, recent studies have investigated the potential of ICL in the visual domain, and pioneering works like MAE-VQGAN \cite{bar2022visual}, SegGPT \cite{wang2023seggpt}, and PIT \cite{fang2024explore} have shown the promise.
Based on their successes, a series of works \cite{zhang2024makes,xu2024towards,zhang2024instruct,sun2023exploring,suo2024rethinking} have extensively focused on improving the effectiveness of Visual ICL (VICL). 
For example, VPR \cite{zhang2024makes} led the way by fine-tuning the visual encoder of CLIP through contrastive learning. 
InMeMo \cite{zhang2024instruct} enhanced the VICL through prompt tuning, alleviating the model’s heavy reliance on high-quality examples. 
Partial2Global \cite{xu2024towards} employed global supervision signals on a novel ranking model to tackle the definition difficulty of positive and negative prompts in contrastive learning. 
In general, existing VICL methods have primarily focused on single prompt selection due to the resolution restrictions of model inputs, while remaining challenges in leveraging multiple prompts. 
Downsampling methods like VPR reduce the resolution of sub-images to meet input requirements, resulting in notable loss of visual details and degraded performance. 
On the other hand, ensemble-based methods like Prompt-SelF \cite{sun2023exploring} develop voting mechanisms that require traversing all prompts and fusing multiple outputs, which rely on task-specific strategies and incur substantial computational overhead. 

In contrast, we effectively address this limitation from a new perspective \emph{collaborative condensation}. 
By condensing multiple prompts at the input level and integrating the most meaningful context from each, we obtain a better-suited prompt for the query image, improving the state of the art while enjoying satisfactory efficiency. 

%% file: sections/Method.tex
\section{Learning Prompt Condensation for VICL}
\label{sec:method}
\subsection{Problem Formulation and Method Overview} \label{subsec:overview}
Suppose there is an training set $\calQ=\dak{I_q,L_q}_{q=1}^{N_q}$, where $I_q,L_q\in\bbR^{\frac{H_0}2\times \frac{W_0}2\times3}$ respectively denote a query image and its ground-truth label, \ie a synthetic image annotating bounding box or segmentation mask. 
Here, $H_0=W_0=224$. 
Additionally, there is an auxiliary task-specific prompt database $\calP=\dak{P_i}_{i=1}^{N_p}$ for contextual reference, where each prompt $P_i=(I_i, L_i)$ is an image-label pair. 
Given a query image $I_q$ sampled from $\calQ$, we obtain a subset of relevant prompts according to image-image similarity, denoted by $\calS_q=\dak{P_{c_k}}_{k=1}^K$. 
For competitive prompt selection, we only choose the rank-1 prompt $P_{c_1}$ to construct contextualized sample, namely
$X_{c_1} = \begin{bmatrix}
        I_{c_1} & L_{c_1} \\
        I_q & L_\texttt{[MASK]}
    \end{bmatrix}\in\bbR^{H_0\times W_0\times3}$. 
Here $L_\texttt{[MASK]}$ denotes the padding mask for the query. 

In contrast, \textbf{prompt condensation} aims to condense $\calS_q$ into a single prompt $P_{c*}$, such that the constructed sample $X_{c*}$ helps the vision backbone $f(\cdot)$ predict the masked query label more accurately than using $X_{c_1}$. 
To achieve this goad, we devise a lightweight plugin module, \modelname{}, denoted as $\phi(\cdot,\cdot)$ and design effective end-to-end learning strategy for optimization. 
\Cref{fig:arc} illustrate the general pipeline. 
Details will be elaborated in the following sections.

\subsection{Preliminaries: MAE-VQGAN as Backbone}\label{subsec:maevqgan}
\begin{figure}[h]
    \centering
    \includegraphics[width=\linewidth]{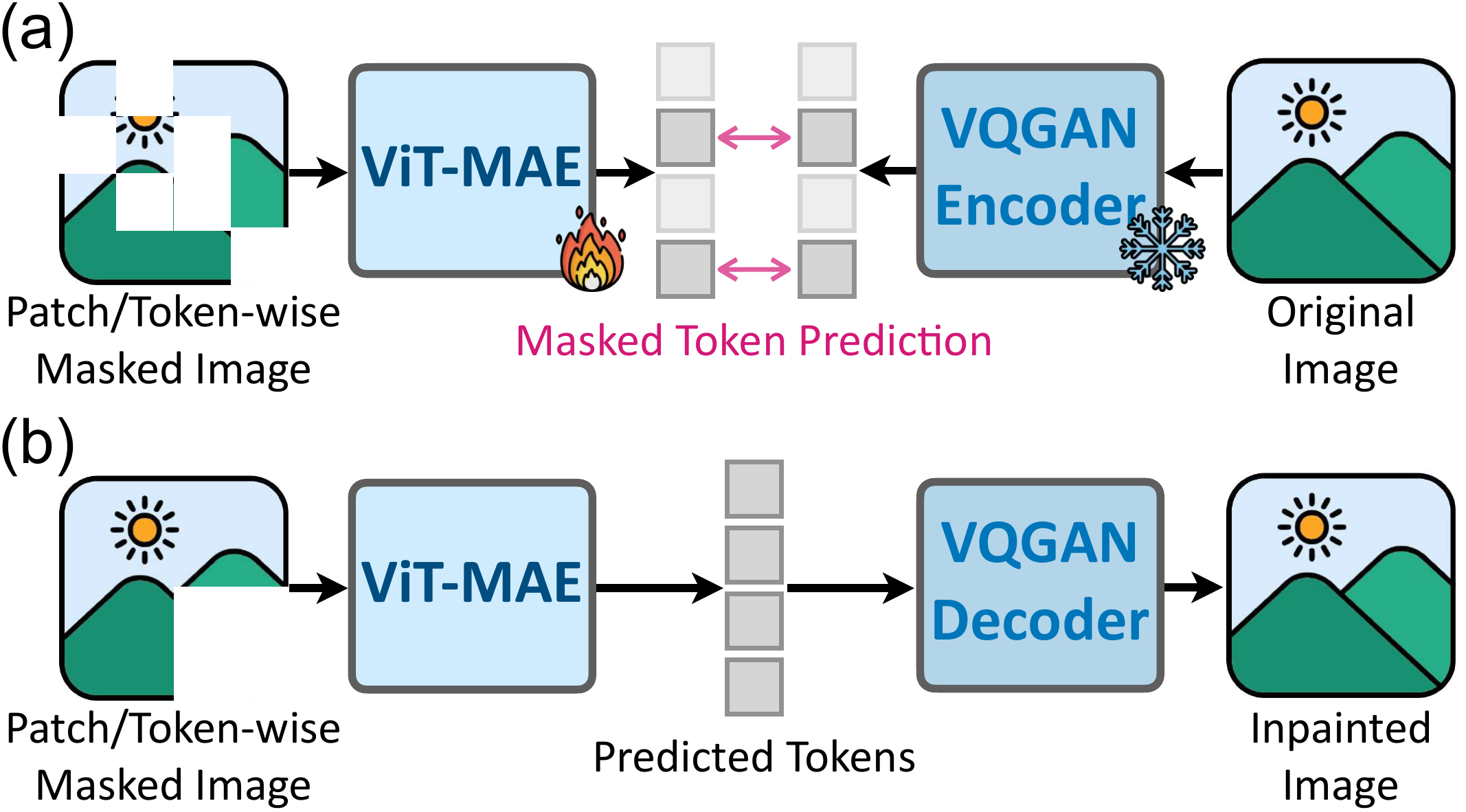}
    \caption{MAE-VQGAN \cite{bar2022visual} neatly combines MAE \cite{he2022masked} and VQGAN \cite{VQGAN_2021} for inpainting. MAE subsequently acts as the backbone in VICL. 
    (\textbf{a}) For pre-training, the MAE is trained to predict the masked tokens. 
    (\textbf{b}) For inference, the VQGAN decoder decodes pixel results from MAE's predicted tokens. 
    }
    \label{fig:maevqgan}
\end{figure}

\citet{bar2022visual} suggested that various vision tasks can be treated as grid inpainting problems and explored the contextual learning capacity of vision models in pixel space. 
They introduced the simple yet powerful MAE-VQGAN model, which served as the foundation for several subsequent works \cite{zhang2024makes,xu2024towards,zhang2024instruct,sun2023exploring} and has established a recognized vision backbone in VICL research.

MAE-VQGAN features an asymmetric input-output architecture: the input is processed through patch embeddings, while the output is generated using the VQGAN decoder. \Cref{fig:maevqgan} outlines the training and inference stages of MAE-VQGAN. For the optimization of \modelname{}, we use the pre-trained MAE-VQGAN checkpoint and keep all its parameters fixed during the process.

\subsection{\textbf{\modelname{}} Design}\label{subsec:condenser}
\begin{figure}[ht]
    \centering
    \includegraphics[width=\linewidth]{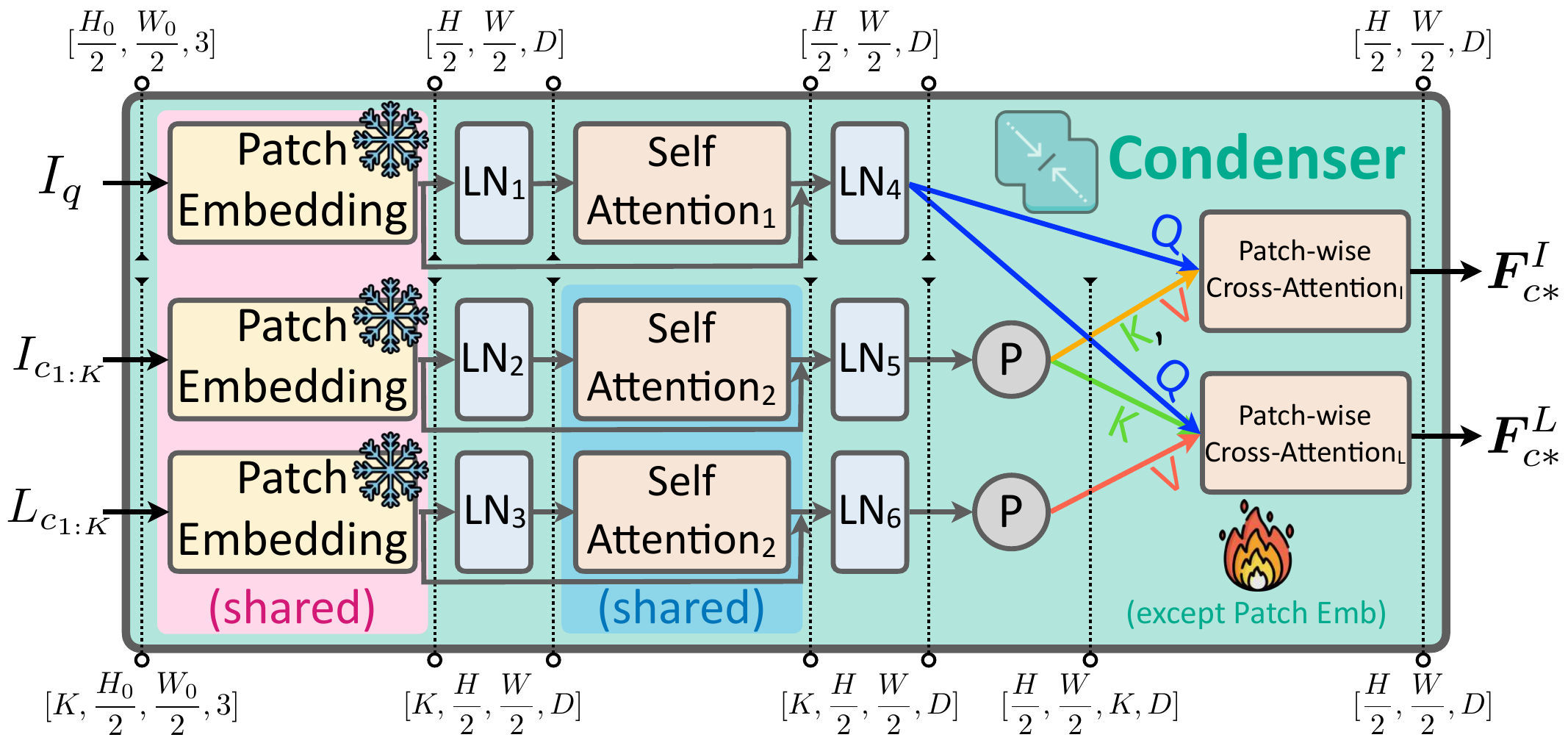}
    \caption{\modelname{} design. 
    The cross-prompt attention performs \emph{patch-wise} cross-attention to aggregate informative context among the $K$ prompts. 
    LN: layer normalization. P: permutation.}
    \label{fig:condenser}
\end{figure}

We design \modelname{} as a lightweight external plugin, expected to capture and integrate relevant fine-grained context from candidate prompts in a query-adaptive manner. 
\Cref{fig:condenser} visualizes the design, which, when combined with the following textual description, will aid in understanding. 

We first embed the query $I_q\in\bbR^{\frac{H_0}2\times \frac{W_0}2\times 3}$ into $\bmF_q^I\in\bbR^{\frac H2\times \frac W2\times D}$, the prompt images $I_{c_{1:K}}\in\bbR^{K\times \frac{H_0}2\times \frac{W_0}2\times 3}$ into $\bmF_{c_{1:K}}^I\in\bbR^{K\times \frac H2\times \frac W2\times D}$, and the stack of prompt $L_{c_{1:K}}\in\bbR^{K\times\frac{H_0}2\times \frac{W_0}2\times 3}$ into $\bmF_{c_{1:K}}^L\in\bbR^{K\times \frac H2\times \frac W2\times D}$, using patch embedding in MAE-VQGAN. Here, $H=W=\frac{H_0}{16}=\frac{W_0}{16}$.

Then, we identify informative patches and patterns within each image with simple self-attention. 
In particular, we share the attention layer between prompt images and their labels to preserve their spatial correspondence. 
The process can be formulated as follows:
\begin{gather}
    \bmF_q^{I(1)}=\text{LN}_\mathrm{4}(\text{SA}_\mathrm{1}(\text{LN}_\mathrm{1}(\bmF_q^I))+\bmF_q^I), \\
    \bmF_{c_{1:K}}^{I(1)}=\text{LN}_\mathrm{5}(\text{SA}_\mathrm{2}(\text{LN}_\mathrm{2}(\bmF_{c_{1:K}}^I))+\bmF_{c_{1:K}}^I), \\
    \bmF_{c_{1:K}}^{L(1)}=\text{LN}_\mathrm{6}(\text{SA}_\mathrm{2}(\text{LN}_\mathrm{3}(\bmF_{c_{1:K}}^L))+\bmF_{c_{1:K}}^L).
\end{gather}
LN and SA denote layer normalization and self-attention, respectively. 
Subscripts differentiate different modules. 
After that, we permute $\bmF_{c_{1:K}}^{I(1)}\in\bbR^{K\times \frac H2\times \frac W2\times D}$ into $\bmF_{c_{1:K}}^{I(2)}\in\bbR^{\frac H2\times \frac W2\times K\times D}$ and
$\bmF_{c_{1:K}}^{L(1)}\in\bbR^{K\times \frac H2\times \frac W2\times D}$ into $\bmF_{c_{1:K}}^{L(2)}\in\bbR^{\frac H2\times \frac W2\times K\times D}$ for later steps. 

Next, we attend and aggregate query-adaptive context with \emph{Patch-wise} Cross-Attention (PCA), namely 
\begin{gather}
    \label{equ:pca_I}
    \bmF_{c*}^I = \text{PCA}_I(\textcolor{myblue}{\bmF_q^{I(1)}}, \textcolor{myorange}{\bmF_{c_{1:K}}^{I(2)}}, \textcolor{mygreen}{\bmF_{c_{1:K}}^{I(2)}}), \\
    \label{equ:pca_L}
    \bmF_{c*}^L = \text{PCA}_L(\textcolor{myblue}{\bmF_q^{I(1)}}, \textcolor{myorange}{\bmF_{c_{1:K}}^{I(2)}}, \textcolor{mygreen}{\bmF_{c_{1:K}}^{L(2)}}).
\end{gather}
PCA$_I(\textcolor{myblue}{\cdot},\textcolor{myorange}{\cdot},\textcolor{mygreen}{\cdot})$ and PCA$_L(\textcolor{myblue}{\cdot},\textcolor{myorange}{\cdot},\textcolor{mygreen}{\cdot})$ condense the context within query images and labels, respectively. 
The 3 arguments of PCAs are \textcolor{myblue}{query}, \textcolor{myorange}{key} and \textcolor{mygreen}{value} factors, respectively. 

For a query image patch, cross-attention to prompt images and labels is restricted to patches in the same spatial position. 
For instance, the cross-attention output for the query patch at position $(h, w)$, where $1 \le h \le \frac H2$ and $1 \le w \le \frac W2$, is
\begin{gather}
    \label{equ:cross-attention-I}
    \bmF_{c*}^I[h,w] = \bmA_{h,w}^I\textcolor{mygreen}{\bmW_V^I\bmF_{c_{1:K}}^{I(2)}[h,w]}, \\
    \label{equ:cross-attention-L}
    \bmF_{c*}^L[h,w] = \bmA_{h,w}^L\textcolor{mygreen}{\bmW_V^L\bmF_{c_{1:K}}^{L(2)}[h,w]}. 
\end{gather}
The attention scores $\bmA_{h,w}^I, \bmA_{h,w}^L$ are with different parameters but the same input to ensure spatial correspondence:
\begin{gather}
    \label{equ:cross-attention-scores-I}
    \resizebox{.9\hsize}{!}{
    $\bmA_{h,w}^I=\text{softmax}\xiaok{\frac{\textcolor{myblue}{(\bmW_Q^I\bmF_q^{I(1)}[h,w])}\textcolor{myorange}{(\bmW_K^I\bmF_{c_{1:K}}^{I(2)}[h,w])}^\top}{\sqrt{D}}}$}, \\
    \label{equ:cross-attention-scores-L}
    \resizebox{.9\hsize}{!}{
    $\bmA_{h,w}^L=\text{softmax}\xiaok{\frac{\textcolor{myblue}{(\bmW_Q^L\bmF_q^{I(1)}[h,w])}\textcolor{myorange}{(\bmW_K^L\bmF_{c_{1:K}}^{I(2)}[h,w])}^\top}{\sqrt{D}}}$}.
\end{gather}
Shapes of parameters and variables in \Cref{equ:cross-attention-I,equ:cross-attention-L,equ:cross-attention-scores-I,equ:cross-attention-scores-L}: 
\begin{gather*}
    \textcolor{myblue}{\bmW_Q^I},\textcolor{myorange}{\bmW_K^I},\textcolor{mygreen}{\bmW_V^I},\textcolor{myblue}{\bmW_Q^L},\textcolor{myorange}{\bmW_K^L},\textcolor{mygreen}{\bmW_V^L}\in\bbR^{D\times D}, \\
    \bmA_{h,w}^I,\,\bmA_{h,w}^L\in\bbR^{K\times K}, \\
    \bmF_q^{I(1)}[h,w],\,\bmF_q^{L(1)}[h,w]\in\bbR^{D}, \\
    \bmF_{c_{1:K}}^{I(2)}[h,w],\,\bmF_{c_{1:K}}^{L(2)}[h,w]\in\bbR^{K\times D}. 
\end{gather*}

Patch-wise cross-attention is preferred over full cross-attention to preserve local consistency in images. 
Restricting attention to the same spatial locations across the query and candidate prompts avoids irrelevant information from unrelated positions. 
This ensures accurate spatial alignment in most tasks with well-defined visual structures, where patch-wise attention provides more precise and consistent results. 

\subsection{Learning Objectives}\label{subsec:objectives}
We integrate the frozen vision backbone into the \modelname{} training process to make them more compatible and acquire more accurate optimization feedback. 
Below we introduce the two learning objectives for optimization.

\subsubsection{Token Prediction for Query Labels}\label{subsubsec:token-prediction}
We first introduce end-to-end token prediction to enable task-oriented supervisory signals. 
As shown in \Cref{fig:arc}, we construct contextualized labeled samples for each candidate prompt in $\calS_q$. 
For example, 
$\tilde X_{c_i} = \begin{bmatrix}
    I_{c_i} & L_{c_i} \\
    I_q & L_q
\end{bmatrix}\in\bbR^{H_0\times W_0\times3}$ for $I_{c_i}$, 
where $L_q$ is the label of $I_q$ for training. 
We then encode $\tilde X_{c_i}$ with the VQGAN encoder into tokens:
\begin{equation}
    \resizebox{.88\hsize}{!}{
$\bmT_{c_i}^{\tilde{X}}=h(\tilde{X}_{c_i})=
\begin{bmatrix}
        \bmT_{c_i}^I & \bmT_{c_i}^L \\
        \bmT_{q_{c_i}}^I & \bmT_{q_{c_i}}^L
    \end{bmatrix}\in\{1,2,\cdots,N_t\}^{H\times W}.$
}
\end{equation}
Here, $h(\cdot)$ represents the VQGAN encoder. $\bmT_{q_{c_i}}^I,\bmT_{q_{c_i}}^L\in\{1,2,\cdots,N_t\}^{\frac H2\times \frac W2}$ are the contextualized encoded tokens for query image and label. $N_t$ is codebook size of VQGAN. 
We slice out $\bmT_{q_{c_i}}^L$ as the labels for token prediction. 

On the backbone side, we obtain the embedded contextualized unlabeled sample as $\bmF^X_{c*}$, predicting output token probabilities through the backbone $f$ as 
\begin{equation}
    \hat{\bmT}_{q_{c*}}^X=
    \begin{bmatrix}
        \hat{\bmT}_{c_*}^I & \hat{\bmT}_{c_*}^L \\
        \hat{\bmT}_{q_{c_*}}^I & \hat{\bmT}_{q_{c_*}}^L
    \end{bmatrix}\in[0,1]^{H\times W\times N_t}.
\end{equation}
We separate query label prediction $\hat{\bmT}_{q_{c*}}^L\in[0,1]^{\frac H2\times \frac W2\times N_t}$ and compute the candidate-specific token prediction loss as 
\begin{equation}
    \ell_\text{TP}^{(i)} = -\frac{4}{HW}\sum_{h}^{H/2}\sum_{w}^{ W/2}\log\hat{\bmT}_{q_{c*}}^L[h,w,\bmT_{q_{c_i}}^L[h,w]].
\end{equation}
The final token prediction loss is summarized by
\begin{equation}\label{equ:token_prediction}
    \calL_\text{TP}=\bbE_{q\sim \calQ}\zhongk{\frac1K\sum_i^K\ell_\text{TP}^{(i)}}.
\end{equation}

\subsubsection{Pre-Alignment for Condensed Prompts}\label{subsubsec:pre-alignment}
To improve stability and calibrate biases, we further introduce a pre-alignment objective that guides \modelname{}'s output to align with the query before backbone inference.
This objective is defined by maximizing cosine similarity: 
\begin{gather}\label{equ:pre_alignment}
    \calL_\text{PA}=-\bbE_{q\sim \calQ}\zhongk{\cos(\bmF_{c*}^I,\bmF_{q}^I)+\cos(\bmF_{c*}^L,\bmF_{q}^L)},
\end{gather}
where instance-wise cosine similarity is defined by
\begin{gather}
    \resizebox{.88\hsize}{!}{
    $\cos(\bmF_{c*}^I,\bmF_{q}^I) = \frac{4}{HW}\sum_h^{H/2}\sum_w^{W/2}\cos(\bmF_{c*}^I[h,w],\bmF_{q}^I[h,w])$}, \\
    \resizebox{.88\hsize}{!}{
    $\cos(\bmF_{c*}^L,\bmF_{q}^L) = \frac{4}{HW}\sum_h^{H/2}\sum_w^{W/2}\cos(\bmF_{c*}^L[h,w],\bmF_{q}^L[h,w])$}.
\end{gather}

Finally, the overall learning objectives is given by
\begin{equation}\label{equ:total_objectives}
    \min_{\phi}\calL_\text{TP} + \lambda\calL_\text{PA}.
\end{equation}
$\phi$ denotes the model parameters.
$\lambda>0$ is a balancing factor for the two objectives. 

%% file: sections/Experiments.tex
\begin{table*}[t]
\centering
\resizebox{0.9\textwidth}{!}{
\begin{tabular}{ccccccccc}
\toprule
& &  \multicolumn{5}{c}{\textbf{Seg. (mIoU ↑)}} &  &   \\
\multirow{-2}{*}{\textbf{Type}} & \multirow{-2}{*}{\textbf{Model}} & \textbf{Fold-0} & \textbf{Fold-1} & \textbf{Fold-2} & \textbf{Fold-3} & \textbf{Mean} & \multirow{-2}{*}{\textbf{Det. (mIoU ↑)}} & \multirow{-2}{*}{\textbf{Col. (MSE ↓)}} \\
\midrule
\midrule
 & Random \cite{bar2022visual} & 28.66 & 30.21 & 27.81 & 23.55 & 27.56 & 25.45 & 0.67 \\
 & UnsupPR \cite{zhang2024makes} & 34.75 & 35.92 & 32.41 & 31.16 & 33.56 & 26.84 & 0.63 \\
 & SupPR \cite{zhang2024makes} & 37.08 & 38.43 & 34.40 & 32.32 & 35.56 & 28.22 & 0.63 \\
 & Prompt-SelF \cite{sun2023exploring} & 35.69 & 38.25 & 35.86 & 33.37 & 35.79 & 28.08 & 0.63 \\
\multirow{-5}{*}{\makecell{Single \\ Prompt\\ Selection}} & Partial2Global \cite{xu2024towards} & 38.81 & 41.54 & 37.25 & 36.01 & 38.40 & 30.66 & 0.58 \\
\midrule
 & Prompt-SelF$_\text{w/ voting}$ \cite{sun2023exploring} & 42.48 & 43.34 & 39.76 & 38.50 & 41.02 & 29.83 & - \\
\multirow{-2}{*}{Voting} & Partial2Global$_\text{w/ voting}$ \cite{xu2024towards} & \emph{43.23} & 45.50 & 41.79 & 40.22 & 42.69 & 32.52 & - \\
\midrule
PEFT & InMeMo \cite{zhang2024instruct} & 41.65 & 47.68 & \emph{42.43} & 40.80 & 43.14 & 43.21 & - \\
\midrule
\rowcolor[HTML]{f5eef8}
 & \modelname{}$_{K=1}$ & 42.13 & \emph{50.31} & 42.20 & \emph{41.90} & \emph{44.14} & \emph{43.22} & \emph{0.56} \\
\rowcolor[HTML]{e8daef}
\multirow{-2}{*}{\makecell{\rowcolor[HTML]{fdedec}Condensation\\\emph{(Ours)}}} & \modelname{}$_{K=16}$ & \textbf{45.53} & \textbf{52.06} & \textbf{44.33} & \textbf{44.58} & \textbf{46.63} & \textbf{44.64} & \textbf{0.54} \\
\hline
\bottomrule
\end{tabular}}
\vspace{-.7em}
\caption{Performance comparison on foreground segmentation (\textbf{Seg.}), single-object detection (\textbf{Det.}), and image colorization (\textbf{Col.}). The best scores are in \textbf{bold} and the second-best scores in \emph{italic}. ${K=1}$ and $16$ represent using a single prompt and 16 prompts, respectively.}
\vspace{-1.5em}
\label{tab:main_table}
\end{table*}

\section{Experiments} \label{sec:experiments}
\subsection{Experimental Setup} \label{subsec:exp_setup}
\subsubsection{Datasets}
Following previous works \cite{zhang2024makes,xu2024towards,bar2022visual}, we conduct experiments on 3 datasets for different tasks: 
(\textbf{i}) \textbf{Pascal-5$^i$} \cite{shaban2017one}, a \emph{foreground segmentation} dataset with 20 categories equally split into 4 folders. 
Each folder contains between 346 and 725 images and associated segmentation masks. For each category, the data contains several image-mask pairs, together with held-out image queries. 
Following \citet{zhang2024instruct}, we use 2286, 3425, 5883, and 2086 in-context samples for training on four folders, respectively. 
(\textbf{ii}) \textbf{Pascal VOC 2012} \cite{everingham2015pascal}, a \emph{single-object detection} dataset with 20 categories. Each sample contains an image and its associated detection box. 
We use 612 in-context samples for training. 
(\textbf{iii}) For the \emph{colorization} task, we use the sampled subsets of \textbf{ImageNet-1K ILSVRC2012} \cite{russakovsky2015imagenet}. 
We randomly sample 50k images from the original 1.2M training set and take the original validation set as our test set. 
We convert each image into the grayscale version to be the task input image while the original image serves as the label.

\subsubsection{Evaluation Metrics}
For performance evaluation, we adopt the standard metrics including \emph{mean Intersection over Union} (\textbf{mIoU}) on foreground segmentation and single-object detection, and regression \emph{Mean Squared Error} (\textbf{MSE}) to assess the performance of image colorization.
For efficiency evaluation (\S \ref{subsubsec:time_and_gpu}, \S \ref{subsubsec:prompt_number_analysis}), we choose to report the average \emph{Inference Time} (in ms) and \emph{GPU Cost} (in MB) per query.

\subsubsection{Implementation Details}
We adopt the pre-trained MAE-VQGAN \cite{bar2022visual} as the backbone and freeze its parameters during training. 
We train and evaluate our model with $K = 1, 2, 4, 8, 16, 32$ candidate prompts accordingly. 
For training, we use the SGD optimizer with a learning rate initialized to 0.03 and decaying according to the cosine schedule. 
We train at most 150 epochs on segmentation and detection, and at most 10 epochs on image colorization. The experiments were conducted on single A100 80G GPUs with a batch size of 16. Unless specified, the loss weight $\lambda$ in \cref{equ:total_objectives} is set to 0.4.

\begin{table}[t]
    \centering
    \small
    \resizebox{1.0\linewidth}{!}{
    \begin{tabular}{c c c c c c}
        \toprule
        \multirow{2}{*}{\textbf{Model}} & \multicolumn{5}{c}{\textbf{Seg. (mIoU ↑)}} \\
        & \textbf{Fold-0} & \textbf{Fold-1} & \textbf{Fold-2} & \textbf{Fold-3} & \textbf{Mean} \\
        \midrule
        \midrule
        InMeMo \cite{zhang2024instruct} & 38.74 & 43.82 & 40.45 & 37.12 & 40.03  \\
        Prompt-SelF \cite{sun2023exploring} & 40.13 & 42.14 & 37.84 & \textbf{38.52} & 39.66  \\
        \rowcolor[HTML]{f5eef8} \modelname{}$_{K=1}$ & \textbf{40.39} & 44.54 & 40.23 & 36.33 & 40.37  \\
        \rowcolor[HTML]{e8daef} \modelname{}$_{K=16}$ & 40.37 & \textbf{44.85} & \textbf{41.03} & 35.84 & \textbf{40.52}  \\
        \hline
        \bottomrule
    \end{tabular}
    }
    \vspace{-.7em}
    \caption{Cross-dataset performance evaluation. We train models on COCO-5$^i$ and test on Pascal-5$^i$ for the segmentation task.}
    \vspace{-1em}
    \label{tab:coco}
\end{table}

\subsection{Comparison with State-of-The-Arts}  \label{subsec:sota_experiments}
\subsubsection{Baselines}
We comprehensively compare
\modelname{} with representative baselines categorized into 3 types: 
(\textbf{i}) \emph{Single Prompt Selection}: including random selection \cite{bar2022visual}, VPR \cite{zhang2024makes} with optimized prompt retrievers, and Partial2Global \cite{xu2024towards} with a hierarchical prompt ranker; 
(\textbf{ii}) \emph{Voting}: Prompt-SelF \cite{sun2023exploring} and Partial2Global \cite{xu2024towards} by output ensemble among different arrangements of the same query-prompt pairs on the input canvas. 
(\textbf{iii}) \emph{PEFT Enhancement}: InMeMo \cite{zhang2024instruct} with pixel instruction tuning for better adapting backbone.

\subsubsection{Performance under Standard Protocols}
As shown in Table \ref{tab:main_table}, our \modelname{} consistently outperforms baselines across the tasks, with further improvement in multi-prompting.
Compared to the strongest baseline, InMeMo, \modelname{} notably achieves relative improvements of 8.09\% and 3.31\% in the segmentation and detection tasks, respectively. 
For the image colorization task, we also outperform the previous top baseline, Partial2Global, with a relative performance improvement of 6.9\%.
These results suggest the effectiveness of the new perspective of \textbf{prompt condensation} in VICL and highlight the advantages of \modelname{}'s in leveraging multiple prompts. 

\begin{table}[t] 
        \centering 
        \small
        \resizebox{0.85\columnwidth}{!}{
        \begin{tabular}{c c c}
            \toprule
            \multirow{2}{*}{\textbf{Method}} & \textbf{Inference Time} & \textbf{GPU Cost} \\
            & \textbf{(ms/query)} & \textbf{(MB/query)} \\
            \midrule
            \midrule
            Retrieval Only \cite{zhang2024makes} & 53.45 & 413.42 \\
            InMeMo \cite{zhang2024instruct} & 55.41 & 497.13 \\
            Prompt-SelF$_{K=16}$ \cite{sun2023exploring} & 989.62 & 446.71 \\
            \rowcolor[HTML]{f5eef8} \modelname{}$_{K=1}$ & 59.17  & 565.42 \\ 
            \rowcolor[HTML]{e8daef} \modelname{}$_{K=16}$ & 66.61 & 1021.86 \\
            \hline
            \bottomrule
        \end{tabular}}
        \vspace{-.7em}
        \caption{Efficiency comparison.}
        \vspace{-1em}
        \label{tab:time_cost_comp_others}
\end{table}

\begin{table*}[t]
\centering
\resizebox{0.8\textwidth}{!}{
\begin{tabular}{cccccccc}
\toprule
& & \multicolumn{5}{c}{\textbf{Seg. (mIoU ↑)}} &\\
\multirow{-2}{*}{\textbf{ID}} &
  \multirow{-2}{*}{\textbf{Model}} &
  \textbf{Fold-0} &
  \textbf{Fold-1} &
  \textbf{Fold-2} &
  \textbf{Fold-3} &
  \textbf{Mean} &
  \multirow{-2}{*}{\textbf{Det. (mIoU ↑)}} \\
\hline
\midrule
\rowcolor[HTML]{f5eef8} (0)  & \modelname{}$_{K=1}$  & 42.13 & 50.31 & 42.20 & 41.90 & 44.14   & 43.22 \\
\rowcolor[HTML]{e8daef} (1)  & \modelname{}$_{K=16}$ & \textbf{45.53} & \textbf{52.06} & \textbf{44.33} & \emph{44.58} & \textbf{46.63} & \textbf{44.64} \\
\midrule
\midrule
(2)  & Mean Pooling            & 18.66 & 18.30  & 16.79 & 15.18 & 17.23 & 17.84 \\
(3)  & w/ Full CA & 44.00 & 49.97 & 42.55 & 42.97 & 44.87 & 43.11 \\
\midrule
\midrule
(4)  & w/o $\mathcal{L}_\text{PA}$ $_{K=1}$                & 42.20 & 49.35 & 42.20 & 41.24 & 43.75 & 43.32 \\
(5)  & w/o $\mathcal{L}_\text{PA}$ $_{K=16}$               & \emph{44.38} & \emph{51.38} & \emph{44.12}  & \textbf{44.77} & \emph{46.16}   & \emph{44.37} \\
(6)  & w/o $\mathcal{L}_\text{TP}$ $_{K=1}$                & 11.50  & 21.46 & 18.70 & 14.87  & 16.63  & 14.91  \\
(7)  & w/o $\mathcal{L}_\text{TP}$ $_{K=16}$               & 8.53 & 9.04  & 8.79  & 8.27  & 8.66  & 6.75  \\
\midrule
\midrule
(8)  & Output Fusion$_{K=16}$ \cite{sun2023exploring}                       & 36.83 & 39.41 & 36.23 & 34.71 & 36.80  & 29.76 \\
\midrule
\midrule
(9) & MAE-VQGAN$_\text{LoRA}$ \cite{hu2021lora}                     & 39.78 & 44.59 & 38.88 & 35.12 & 39.49 & 36.66  \\
(10) & MAE-VQGAN$_\text{P-Tuning V2}$ \cite{p_tuning_v2}                 & 41.37 & 46.72  & 42.45 & 38.41 & 42.24 & 39.07 \\
\hline
\bottomrule
\end{tabular}
}
\vspace{-.7em}
\caption{Ablation study of \modelname{}. The best scores are marked in \textbf{bold} and the second-best scores in \emph{italic}.}
\vspace{-1em}
\label{tab:abla}
\end{table*}

\subsubsection{Cross-Dataset Transferability}
In practice, generalizability is an essential requirement for VICL since models are expected to adapt to diverse scenes through reasoning in context. 
Here we compare \modelname{} with strong competitors in this regard by evaluating cross-dataset performance. 
Following InMeMo \cite{zhang2024instruct} and Prompt-SelF \cite{sun2023exploring}, we construct a COCO-5$^i$ dataset selected from MSCOCO \cite{lin2014microsoft} that matches the same categories in Pascal-5$^i$. 
Then we train the models on COCO-5$^i$ and test them on Pascal-5$^i$. 
Table \ref{tab:coco} reports the results, where we can learn the favorable downstream performance of \modelname{}. 
In particular, condensing multiple prompts shows remarkable advantages over single-prompting. 
We attribute the success to the collaboration of different prompts that individually contribute important fine-grained cues, which thereby enhances prediction against noises and domain gaps. 

\subsubsection{Inference Time and GPU Overhead}\label{subsubsec:time_and_gpu}
As \modelname{} employs multiple prompts, efficiency is a key consideration. 
We compare the computational and memory overheads of existing methods, including:
(\textbf{i}) Retrieval-only methods, such as those of `single prompt selection' in \Cref{tab:main_table}, which introduce retrieval overhead to the task inference. 
(\textbf{ii}) InMeMo \cite{zhang2024instruct} with more extra overhead for adding pixel prompt. 
(\textbf{iii}) Prompt-SelF \cite{sun2023exploring} with a refined retrieval process and output ensemble strategy that duplicates the forward steps for the backbone. 
We present the result in \Cref{tab:time_cost_comp_others}. 
It is evident that Prompt-SelF severely increases inference time when using multiple prompts and thus falls short of efficient usage. 
In contrast, our \modelname{} preferably manages extra computational burden compared to single-prompting methods, maintaining a satisfactory inference throughput. 
While \modelname{} does introduce more memory overhead than the baselines, this is still within an acceptable range and will be relieved with larger backbones.

\subsection{Model Ablation}  \label{subsec:ablation_study}
In this section, we investigate the efficacy of different components to facilitate a better understanding of our method. 
For ease of reference and comparison, we assign IDs to different ablated model variants. 
For example, we associate Variants (0) and (1) with our default models, as shown in \Cref{tab:abla}.

\subsubsection{Effectiveness of \modelname{} Design}
To investigate the \modelname{} design, we first construct the following variants: 
(\textbf{2}) removes the \modelname{} and uses mean pooling to fuse multiple prompts; 
(\textbf{3}) replaces the Patch-wise Cross-Attention (PCA) module in \cref{equ:pca_I,equ:pca_L} with the full attention cross-attention that allows a query patch to attend to any patch position of the prompts. 
We then experiment with 16 candidate prompts and report the results in \Cref{tab:abla}.  

By comparing Variants (1) and (2), we can see that prompt condensation is rather than simply pooling the context across prompts, indicating its significant efficacy in extracting and coordinating contexts in raw prompts. 
By comparing Variants (1) and (3), we can also observe that enabling full cross-attention is even inferior to directly removing query-prompt interactions, revealing the non-trivial design of PCA that preserves the local consistency in the condensed prompt.
This is consistent with our intuition (See the end of \S \ref{subsec:condenser}).

\vspace{-.5em}
\subsubsection{Effects of Learning Objectives}
We construct the following variants to study the contributions of the 2 learning objectives: 
(\textbf{4}) and (\textbf{5}) remove the pre-alignment loss (\cref{equ:pre_alignment}) under single- and multi-prompting settings, respectively;
(\textbf{6}) and (\textbf{7}) remove the token prediction loss (\cref{equ:token_prediction}) under single- and multi-prompting settings. 

By comparing Variants (4) and (5) with the full versions, we see the consistent advantages of adding $\calL_\text{PA}$. 
On the other hand, Variants (6) and (7), which omit the token prediction objective, show significant performance degradation. This suggests that end-to-end backbone feedback is crucial for robust prompt condensation. The poor performance likely stems from \modelname{} operating in the low-level feature space, where the lightweight module struggles to capture comprehensive semantics and reconstruct full details in the labeled query. Thus, we conclude that pre-alignment serves as a regularization objective rather than a complete learning signal for potential \emph{zero-shot} or \emph{backbone-agnostic} training.

\subsubsection{\modelname{} Vs. Output Fusion}
To compare \emph{early fusion} as \modelname{} and \emph{late fusion} as output ensemble, we designed Variant (\textbf{8}) following the voting scheme in \citet{sun2023exploring}. 
We report its performance with 16 prompts in \Cref{tab:abla}, which is relatively inferior to our default design, and neither matches the results of Prompt-SelF with voting using a single query-prompt pair (see \Cref{tab:main_table}). 
This suggests the difficulty in designing multi-prompting strategies via heuristics. 
Fortunately, our end-to-end learned prompt condensation strategy supplies a promising remedy. 

\subsubsection{\modelname{} Vs. PEFT}
We also design PEFT Variants: (\textbf{9}) applies LoRA \cite{hu2021lora} to the backbone, while (\textbf{10}) adopts P-Tuning V2 \cite{p_tuning_v2}, both with aligned parameter size to \modelname{}. 
As shown in Table \ref{tab:abla}, \modelname{} exhibits clear advantages over these variants. 
Moreover, existing PEFT methods are hard to benefit from multi-prompting, 
while \modelname{} breaks this limitation. 

\subsubsection{Impact of Candidate Prompt Number, $K$}\label{subsubsec:prompt_number_analysis}
\begin{figure}[t]
    \vspace{-1em}
    \centering
    \includegraphics[width=0.9\columnwidth]{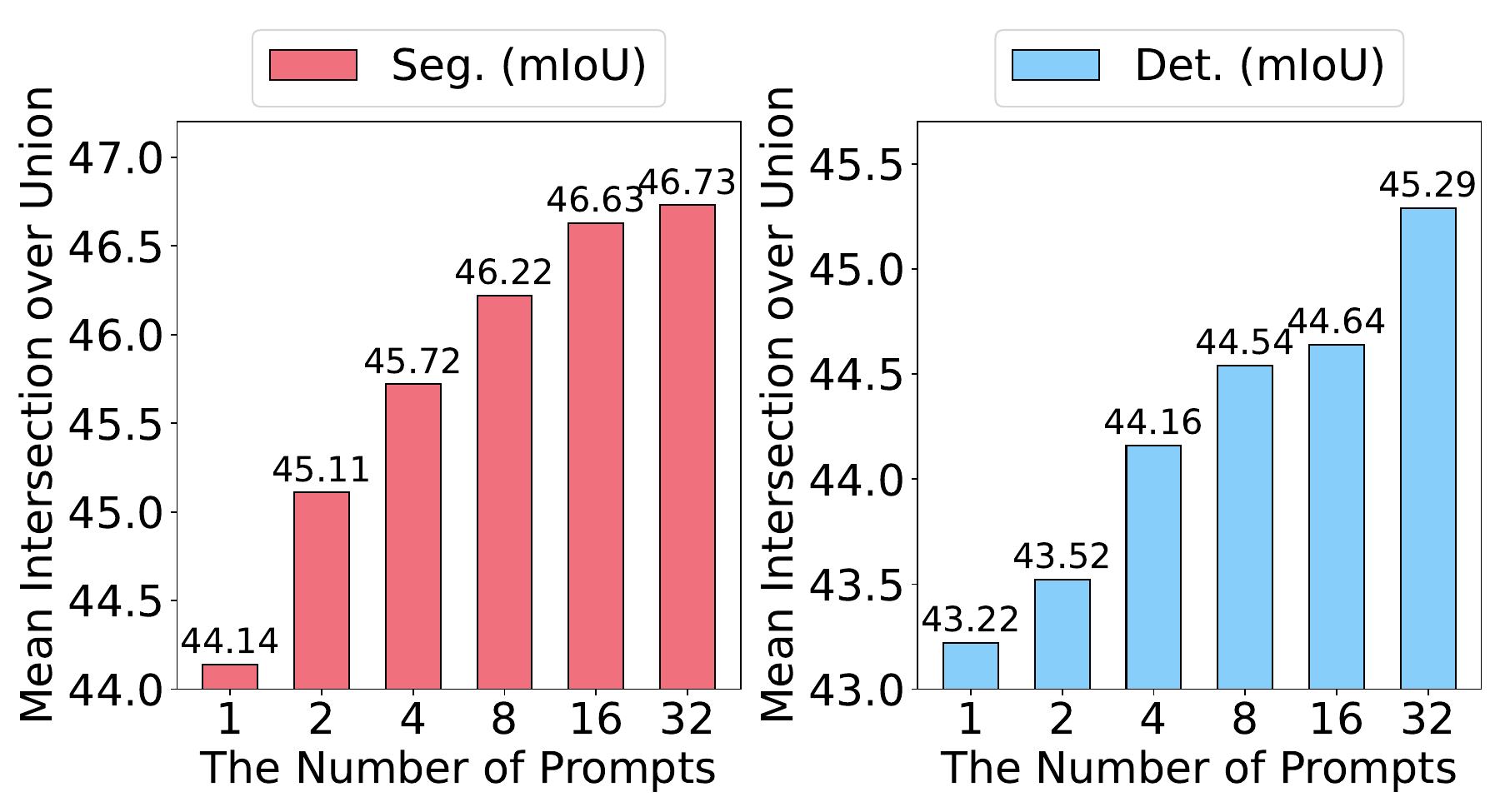}
    \vspace{-1em}
    \caption{Impact of the prompt number $K$ on foreground segmentation, single object detection.}
\label{fig:diff_k}
\end{figure}

\textbf{Impact on Performance Metrics for Various Tasks.} We evaluate \modelname{} with $K = 1, 2, 4, 8, 16, 32$. As shown in Figure \ref{fig:diff_k}, it can be observed that \modelname{} improves progressively with the increase in the number of prompts. This confirms the scalability of our \modelname{}.

\noindent\textbf{Impact on Time Overhead and GPU Cost.} 
As shown in Table \ref{tab:time_and_memory}, \modelname{} exhibits moderate changes in inference time as the value of $K$ varies. In contrast, for the multi-prompt scenario of Prompt-SelF \cite{sun2023exploring}, the time cost increases approximately linearly with $K$, which could hinder practical deployment of the model. Our method also incurs relatively low GPU consumption, with an additional 30MB of GPU memory required for each extra prompt, a cost that can be considered acceptable. This comparison further highlights the efficiency of \modelname{}.

\begin{table}[t]
    \centering
    \small
    \resizebox{\columnwidth}{!}{
    \begin{tabular}{c c c c}
        \toprule
        Prompt &  Inference Time & Inference Time & GPU Cost\\
        Number&for Prompt-SelF & for \modelname{} & for \modelname{}\\
        $K$ & (ms/query) & (ms/query) & (MB/query) \\
        \midrule
        \midrule
        1  & 53.45 & 59.17 & 565.42 \\
        2  & 113.18 & 61.68 & 615.14 \\
        4  & 231.63 & 62.34 & 672.76 \\
        8  & 476.44 & 63.23 & 744.67 \\
        16 & 989.62 & 66.61 & 1021.86 \\
        32 & 1945.15 & 74.13 & 1389.91 \\
        \hline
        \bottomrule
    \end{tabular}
    }
    \caption{Efficiency evaluation under varying prompt numbers. \modelname{} with prompt condensation is more scalable than Prompt-SelF \cite{sun2023exploring} with output fusion.}
    \label{tab:time_and_memory}
\end{table}

\subsubsection{Robustness with Different Retrieval Methods}\label{subsec:different_retrieval}

\begin{table}[t]
\resizebox{1.0\linewidth}{!}{
\begin{tabular}{ccccccc}
\toprule
\multirow{2}{*}{\makecell{\textbf{Retrieval}\\\textbf{Method}}} & \multirow{2}{*}{$K$} & \multicolumn{5}{c}{\textbf{Seg. (mIou ↑)}}   \\
& & \textbf{Fold-0} & \textbf{Fold-1} & \textbf{Fold-2} & \textbf{Fold-3} & \textbf{Mean}    \\
\midrule \midrule
Random                         & 1 & 41.03  & 47.90  & 41.40  & 38.95  & 42.32 \\
UnsupPR \cite{zhang2024makes}                  & 1 & 42.14  & 48.59  & 42.55   & 41.85  & 43.78  \\
SupPR \cite{zhang2024makes}                         & 1 & 41.66   & 49.50  & 42.21  & 41.82  & 43.80 \\
\rowcolor[HTML]{f5eef8}
Pixel-Level Retrieval  \cite{sun2023exploring}  & 1 & 42.13  & 50.31  & 42.20  & 41.90  & 44.14   \\
\midrule
\midrule
Random                         & 16 & 41.68  & 49.20  & 42.80  & 39.37  & 43.26 \\
UnsupPR \cite{zhang2024makes}                  & 16 & 43.35  & 50.39  & 43.47  & 43.44  & 45.16  \\
SupPR \cite{zhang2024makes}                        & 16 & 43.42   & 50.23  & 42.90  & 42.79  & 44.84   \\
\rowcolor[HTML]{e8daef}
Pixel-Level Retrieval \cite{sun2023exploring} & 16 & 45.53  & 52.06  & 44.33  & 44.58  & 46.63 \\
\hline
\bottomrule
\end{tabular}
}
\caption{Performance evaluation of \modelname{} regarding different retrieval methods and prompt numbers.}
\label{tab:different_retrieval}
\end{table}

Following \citet{zhang2024instruct}, we use pixel-level retrieval \cite{sun2023exploring} as the default setting. 
Additionally, we train \modelname{} on segmentation tasks using random retrieval, UnsupPR \cite{zhang2024makes} (\ie pre-trained and fixed CLIP-ViT), and SupPR \cite{zhang2024makes}.
The results are shown in \Cref{tab:different_retrieval}.
We observe that the capability of the retrieval method affects the performance of \modelname{}. 
More powerful retrieval methods provide more accurate raw candidate prompts, which help \modelname{} generate more effective final prompts. 
Fortunately, compared to the results in Table 1, we find that \modelname{} can effectively enhance the robustness of VICL, mitigating performance degradation caused by weaker retrieval methods. 
Notably, in the case of multi-prompting, the performance gap caused by different retrieval methods is further reduced, highlighting the importance of prompt condensation.

%% file: sections/Conclusions.tex
\section{Conclusions}
\label{sec:conclusion}

In this paper, we investigate VICL and reveal that current methods often struggle to select a single prompt from candidates. 
In contrast, we propose collaborating and condensing multiple prompts to foster better context acquisition and generalization. 
We design a lightweight external plugin \modelname{} to capture and integrate fine-grained context, which is optimized end-to-end with the backbone.
We conduct comprehensive experiments and showcase \modelname{}'s strengths, including effective context compression, scalability with more prompts, and computational efficiency, suggesting a favorable solution for VICL. 
We hope our proposed perspective and preliminary results will encourage more work on effective and efficient VICL.

\paragraph{Acknowledgments}
We thank the anonymous reviewers and chairs for their efforts and constructive suggestions. 
This work is supported in part by the National Natural Science Foundation of China under grant 624B2088, 62171248, 62301189, 62302309, the PCNL KEY project (PCL2023AS6-1), and Shenzhen Science and Technology Program under Grant KJZD20240903103702004, JCYJ20220818101012025, RCBS20221008093124061, GXWD20220811172936001.
Long Chen was supported by the Hong Kong SAR RGC Early Career Scheme (26208924), the National Natural Science Foundation of China Young Scholar Fund (62402408).